# Design and Validation of a Lightweight, Low-Profile Powered Knee Prosthesis with Quasi-Direct Drive Actuation

Ross J. Cortino, Ryan Posh, Emily G. Keller, and Robert D. Gregg

***Abstract*—Fully-powered knee prostheses, unlike traditional passive knees, can perform controlled positive work, reducing the need for compensatory behaviors by users during energy-intensive activities. While quasi-direct drive (QDD) actuators provide superior torque control, backdrivability, and acoustic noise properties compared to traditional highly-geared actuators, prior QDD prototypes have been too heavy and bulky for commercial translation. In this work, we present the design and validation of a new lightweight (∼2.6 kg) and low-profile (24.5 cm tip-to-tip build height) QDD knee prosthesis. By optimizing an 18:1 two-stage transmission alongside thermal and structural finite-element analyses, we significantly reduce device mass while enabling a peak torque of 145 Nm. Through benchtop tests, we validate the device's high output torque, low backdrive torque (∼1 Nm), and its precision position and torque control capabilities. We also demonstrate biomimetic kinematics and peak knee extension torques (within one standard deviation of able-bodied references) during both level-ground walking and sit-stand transitions performed by three participants with transfemoral amputation and varying K-levels. By meeting or improving upon the mass, build height, peak torque, and acoustic noise of a leading commercial powered knee, this work establishes the clinical viability of emerging QDD prostheses that promise improved dynamic performance for their users.**

## I. Introduction

Each year millions of individuals worldwide experience limb loss, with 83% of cases in the USA involving a lower limb [1]. Conventional microprocessor-controlled prosthetic knees can restore mobility through their lightweight, adaptive designs, allowing users to perform most activities of daily living (ADLs). However, these quasi-passive designs cannot provide the net-positive work needed for biomimetic ramp or stair ascent and sit-to-stand transitions [2]. Instead, users rely on compensatory behaviors [3] that strain their intact joints, often leading to chronic back pain and osteoarthritis [4]–[6]. Individuals with transfemoral (TF) amputation also can experience increased cognitive load, fall-related incidents, and asymmetric gait with these devices [7], [8].

Through active joint control, powered knee prostheses can perform biomimetic levels of controlled positive and negative work to reduce strain and compensations [9]–[12]. This capability has also been shown to improve inter-limb symmetry [13], [14] and reduce peak hip flexion moments and intact-joint work during locomotion [12], [14]. Despite these biomechanical benefits, the extra weight of powered prosthetic legs can increase ipsilateral hip extension moments and intact-side ankle work compared to using a conventional prosthesis during level-ground walking [12].

Designing lightweight devices that are powerful enough to assist the primary ADLs remains a significant engineering challenge and a barrier to widespread adoption of powered prosthetic legs. Notable academic and commercial endeavors have explored the tradeoff between power and weight through the development of fully-powered (actuator replicates complete muscle function), hybrid-powered (combines fully-powered, quasi-passive, and passive elements), and semi-powered devices (couples quasi-passive or passive elements with a small motor for supplemental torque and swing control) [15], [16]. Semi-powered knee prostheses [17]–[20] are often lighter than fully-powered and hybrid-powered devices and tend to have the added benefit of continued operation during loss of active power, improving user safety and trust. However, their low positive power output fails to satisfy all demands of high-energy tasks like sit-to-stand and stairs, requiring similar compensatory behaviors seen with quasi-passive devices.

To achieve the peak torque demands of the knee (∼120-145 Nm for a 102 kg user [21], [22]) with acceptable size and weight, many fully-powered and hybrid-powered devices utilize a small electric motor paired with a large transmission ratio. Notable fully-powered knees include the belt-driven designs of the Vanderbilt Leg (176:1, ∼2.50 kg knee module) [23] and Open-Source Leg V2 (41.5:1, ∼2.70 kg knee module) [24], the variable transmission design of the previous generation Utah Knee (25-375:1, 1.59 kg) [25], and the series elastic design of the commercial Össur Power Knee (high-ratio harmonic drive, 2.70 kg) [26]. While these devices can provide biomimetic torque and power for most ADLs, their highly-geared transmissions produce substantial acoustic noise and contribute drivetrain dynamics (e.g., reflected inertia/damping) that complicate torque or impedance control [24] and make the joint mechanically rigid. The series-elastic transmissions used in the Össur Power Knee and hybrid-powered knees in [27], [28] provide compliance for impact mitigation, torque sensing for control, and energy storage and release. However, series-elastic actuators (SEAs) tend to have limited control bandwidth and torque output along with greater mechanical complexity.

Other hybrid-powered knee devices created by the Utah HGN Laboratory [29]–[31] pair smaller electric actuators with spring-damper and spring elements to achieve biomimetic torque levels across various ADLs with a lightweight design. The state-of-the-art, hybrid-powered Direct Ball Screw Drive Knee (∼1.9 kg) [31] outputs up to 145 Nm passive torque during eccentric tasks and 115 Nm active torque during concentric tasks. Its force-sensitive transmission provides low output impedance and high backdrivability in low-load con-

This work was supported by the National Institute of Child Health & Human Development (NICHD) of the NIH under Award Number R01HD094772 and by the National Science Foundation under Award Number 2024237. Funding for R. Posh was provided by NICHD under Award Number 1F32HD116414-01A1. The content is solely the responsibility of the authors and does not necessarily represent the official views of the NIH or NSF.

R. J. Cortino, R. Posh, E. G. Keller, and R. D. Gregg are with the Robotics Department, University of Michigan, Ann Arbor, MI 48109. Contact: {cortinrj, rdgregg}@umich.edu

ditions such as swing phase, and the device can be utilized as a passive knee prosthesis for level-ground walking in the event of electrical power loss. Despite these benefits, the variable transmission has two primary limitations: its kinematic singularity at 88° causes torque degradation at high flexion angles encountered during sit-stand transitions, and its output impedance increases with force output during stance phase. The passive elements within the actuator also increase design and manufacturing complexity in a similar manner to SEAs.

To reduce actuator impedance and improve device controllability while maintaining the ability to output biomimetic torque levels across ADLs, other groups have developed fully-powered knee prostheses with high-torque motors paired with constant low-gear-ratio transmissions (≤24:1), known as quasi-direct drive (QDD) actuators. These low-impedance actuation schemes take inspiration from those originally deployed in biped and quadruped robots [32]–[35] and in powered orthoses [36], which have demonstrated benefits including impact mitigation, backdrivable dynamics, high control bandwidth, accurate open-loop torque/impedance control, and reduced acoustic noise. These benefits were verified in the first QDD prosthetic leg [37] using a 22:1 compound planetary gearset to achieve a 182 Nm peak output torque and less than 3 Nm backdrive torque at the knee and ankle. However, these benefits were partially negated by its distal knee center of rotation (CoR), causing misalignment with the intact-side knee center; tall build height, excluding shorter users; and heavy weight (∼3.5 kg for the knee, ∼6.5 kg total with ankle). This excess mass caused effort penalties at the prosthetic-side hip and intact-side ankle during level-ground walking [12] as well as discomfort for tasks that require prolonged leg lifting or constrained foot placement, e.g., stair ascent/descent [38]. A similar knee design with a 36:1 transmission achieved a slightly reduced weight of 3.3 kg [39] with the tradeoff of higher reflected inertia at the joint. Despite the performance benefits of these prior designs, the commercial viability of fully-powered QDD prostheses has yet to be demonstrated.

This work addresses these barriers to clinical translation by introducing a new lightweight, compact, fully-powered knee prosthesis with a highly optimized QDD actuator. The specific technical and clinical contributions of this work are threefold:

1) A Novel QDD Architecture: We present the electromechanical design of a fully-powered QDD knee prosthesis with mass (2.60 kg with batteries) and build-height (24.5 cm) characteristics that meet or improve upon the standards of the commercial Össur Power Knee. Informed by rigorous thermal and structural modeling, we optimize a custom BLDC motor and 18:1 two-stage transmission to overcome the historic mass and size penalties of QDD designs while maintaining full biomimetic capability (145 Nm peak active torque, 81% higher than Power Knee).
2) State-of-the-Art Dynamic Performance: Through benchtop experiments, we demonstrate key benefits of the low-impedance actuator design—a state-of-the-art combination of active output torque, backdrivability, closed-loop position tracking, and open-loop torque tracking.
3) Clinical Feasibility and Usability: We validate the device's real-world clinical viability across multiple participants with transfemoral amputation (varying K-levels, genders, and anthropometrics) during level-ground walking and sit-stand transitions, alongside a comparative demonstration of its reduced acoustic noise profile against the commercial standard, the Össur Power Knee.

This paper is organized in the following manner. Section II details the mechatronic design of the knee prosthesis, including design requirements, motor selection, transmission ratio selections and design optimization, electrical system, and the knee structure. Section III presents the benchtop and human subject validations of the device. Section IV discusses this work in the context of the literature, subject participant feedback, and limitations and future work. Section V concludes this paper.

TABLE I
KNEE PROSTHESIS DESIGN SPECIFICATIONS (UP TO 102 KG USER MASS)

| | Requirements | Achieved |
|---|---|---|
| Peak Torque (Nm) | 145 | 145 |
| Peak Velocity (°/s) | 505 | 916 |
| Peak Power (W) | 420 | 698 |
| Reflected Rotor Inertia (kg·m²) | ≤ 0.038 | 0.037 |
| Mass (kg) | ≤ 2.70 | 2.60 |
| Tip-to-tip Build Height (cm) | ≤ 30.8 | 24.5 |
| CoR Distance (mm) | ≤ 67 | 63.59 |

## II. HARDWARE DESIGN

### A. Design Overview

The goal of this prosthetic knee design was to achieve human-like joint impedance and dynamics across the primary ADLs, with a commercially viable weight, packaging factor, and joint CoR. Table I shows our design requirements, where biomechanical specifications are based on able-bodied (AB) data for a 102 kg subject (75th percentile male weight) [22], [40]–[43]. Human-like joint impedance was specified through the reflected motor inertia, which was limited to 10% of the pendulum inertia of the biological shank and foot (i.e., $0.1 \cdot 0.38$ kg·m²) as done in [20], [31]. The mass limit of ≤ 2.7 kg and tip-to-tip build height limit of ≤ 30.8 cm were based on the Össur Power Knee PKA01 (directly measured), which represents the state-of-the-art in commercial powered knee prostheses. This mass target is 0.8 kg lower than our previous QDD prototype [37]. Furthermore, the geometry of the Open-Source Leg v2 informed our target knee center of rotation (CoR) distance of ≤ 67 mm (measured from the top of male pyramid adapter).

The final assembly in Fig. 1 has favorable mass and dimension quantities compared to the Power Knee [26] and our prior QDD design [37] (all directly measured). The device has a build height of ∼24.5 cm from the proximal to distal pyramid tips (including the load cell assembly), an anterior/posterior depth of 9.94 cm, and a medial/lateral width of 9.46 cm (including externally mounted electronics). Compared to the Power Knee, our device is 6.3 cm shorter, 0.14 cm slimmer in medial/lateral width, and 0.54 cm larger in anterior/posterior depth. Our device weighs 2.60 kg with batteries, successfully achieving our weight goal by coming in 100 grams lighter than the current-generation Power Knee and 900 grams lighter than our prior QDD design. While our device's CoR distance of 63.59 mm is larger than the Power Knee's distance of 52 mm, it is smaller than the Open-Source Leg (67 mm) and substantially smaller than our prior QDD design (80 mm). Finally, the new 18:1 QDD actuator (presented next) has an estimated rotor reflected inertia of 0.037 kg·m² and a calculated actuator inertia of 0.041 kg·m² based on the known geometry and mass of the components (compared to 0.0557 kg·m² and 0.0625 kg·m², respectively, in our prior design [37]).

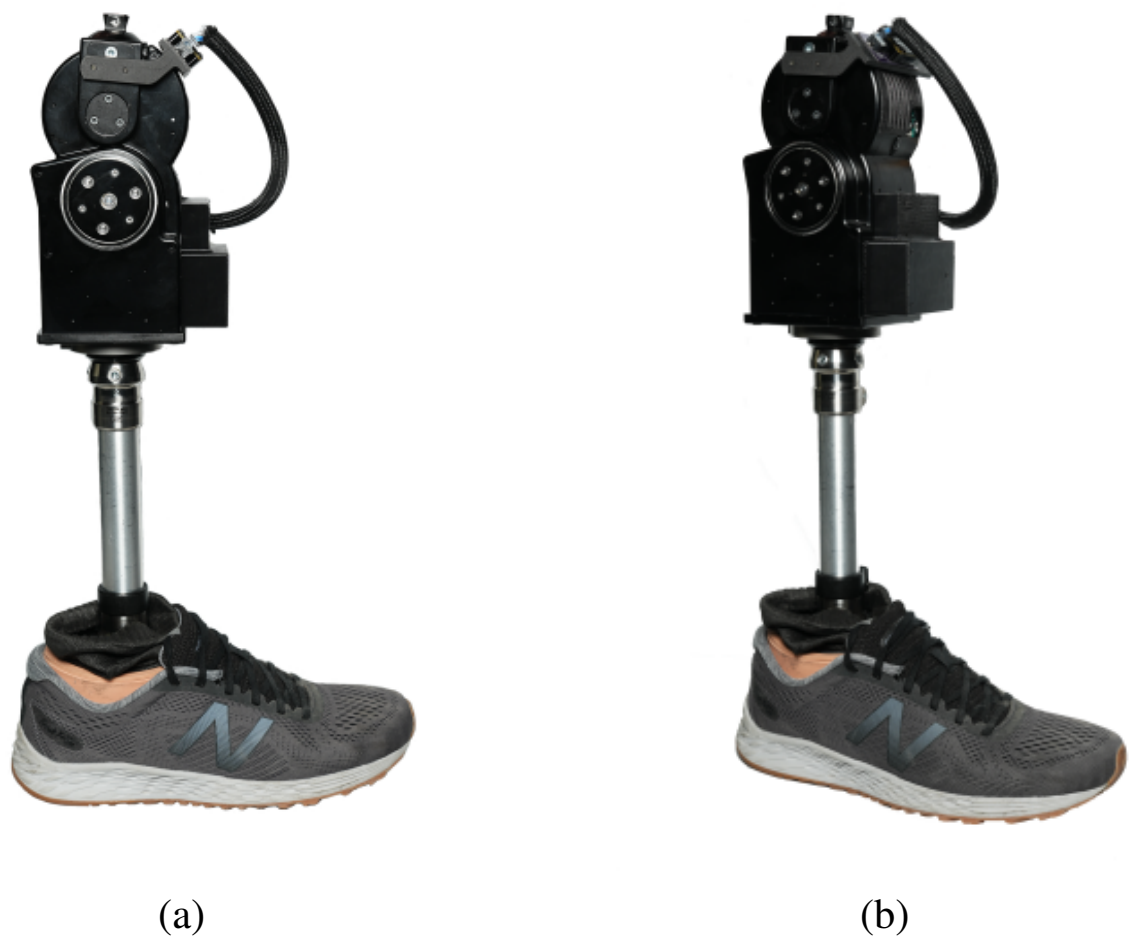

(a) (b)

Fig. 1. Figures (a) and (b) show the final knee prosthesis attached to a passive ankle prosthesis and shoe through a distal pylon. The visible wire sheath connects the top-mounted Microstrain IMU, which can be removed in favor of an internal shank-mounted IMU.

### *B. Motor and Driver*

After determining that current commercial motors were sub-optimal for our requirements, we specified a custom frameless, internal rotor motor from CubeMars (Nanchang, Jiangxi, China) as a lighter alternative to the Robo-Drive ILM 85x26 (TQ-Systems GmbH, Seefeld, Germany) used in our previous prototype [37], achieving a mass reduction of $\sim$33 grams. This RI8523 motor has a manufacturer-rated (power-invariant) torque constant of 0.169 Nm/A with a manufacturer-rated continuous torque of 2.60 Nm, peak torque of 7.8 Nm, and no-load speed of 2750 RPM at 48 V. As described in Section SIII-A, we experimentally determined a (power-invariant) torque constant of 0.175 Nm/A.

We selected the R80/80 Gold Solo Twitter (Elmo Motion Control, Petach-Tikva, Israel) as our motor driver due to its lightweight and compact design, providing a rated continuous current of 56.5 A and a peak current of 80 A. The driver's on-board CAN bus reduces wiring complexity (see Section II-D) and will facilitate seamless integration with future powered ankle designs (see Section IV-C).

### *C. Transmission*

*1) Determining a Transmission Ratio:* The transmission ratio ($i_{\text{TR}}$) is a critical design variable, as it directly governs motor temperature, backdrivability, and reflected inertia at the prosthesis joint. The transmission ratio also re-scales the torque-velocity boundary of the actuator. We chose the minimum necessary ratio ($i_{\text{TR}}$ = 18:1) to contain all AB joint torque-velocity trajectories within the envelope shown in Fig. 2a. The x-intercept is the no-load speed calculated from our motor's rated voltage (48 V) and the torque constant found during our characterization in Supplemental Section SIII-A, and the y-intercept of this boundary represents the motor stall torque scaled by $i_{\text{TR}}$. To reach our desired peak torque of 145 Nm and fit all AB trajectories within the envelope, the motor must be able to repeatedly reach a peak torque of approximately 8.06 Nm. Although this requirement is slightly larger than the manufacturer-rated peak torque (7.8 Nm), this represents a thermal limitation which can be mitigated by enhancing the motor's thermal dissipation under load.

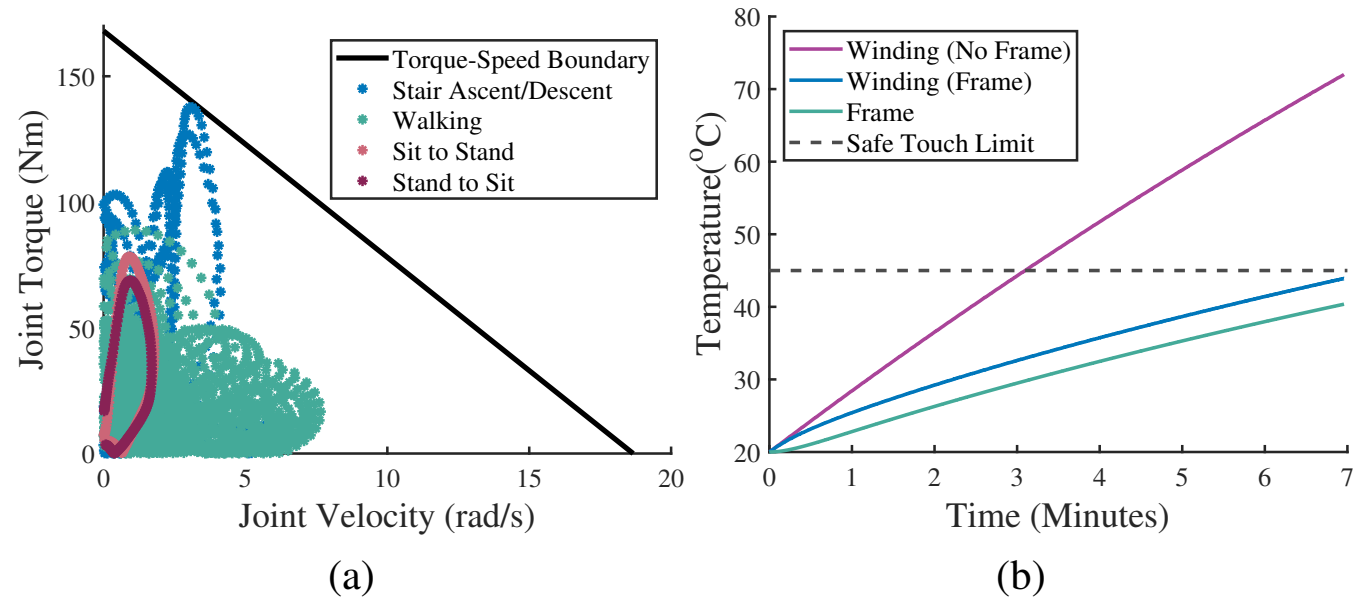

(a) (b)

Fig. 2. (a) Simulated torque-velocity relationship for the RI8523 motor coupled with the chosen 18:1 transmission ratio. (b) Simulated motor winding temperature during 600 stair ascent strides.

Fig. 2b shows the estimated winding temperature for the RI8523 motor and an 18:1 transmission ratio when performing 600 stair ascent strides for a 102 kg user. The stair test case was based on similar thermal modeling work for a powered knee prosthesis in [29]. Across all simulations, we observed that the transmission ratio is inversely correlated with the motor's winding temperature. To enable safe temperatures with $i_{\text{TR}}$ = 18:1, we designed our motor frame to act as a heat sink (see Supplemental Section SIII-B2), thus reducing the steady-state winding temperature.

*2) Transmission Design Optimization:* With the design target of an 18:1 transmission ratio, we chose a two-stage gearbox design where the secondary stage is a simple gear train moving the joint center of rotation proximally towards the user. For the first stage, we desired a compact solution that had a large mechanical advantage relative to its packaging volume. A traditional planetary gear stage was considered, however, they are usually limited to a $i_{\text{TR}}$ of 10:1 due to geometry and are often not the most compact solution. Cycloidal drives and harmonic drives were also considered, however, these tend to be utilized for $i_{\text{TR}}$ greater than or equal to 30:1, which is out of the quasi-direct drive regime. Our previous leg ($i_{\text{TR}}$ = 22:1 in [37]) utilized a single-stage stepped-planet compound planetary gear transmission (SPC-PGT), which can achieve a much higher $i_{\text{TR}}$ within the same geometric volume as a traditional planetary gear set. For similar reasons, we chose this compact gear style for the first stage of our transmission.

The SPC-PGT in this design is constructed of four different gear sizes, with a total of eight gears in the gear set (see Fig. 3a). The sun gear is driven by our motor, meshing with three planet gears spaced 120° radially. The transitional gears are mounted coaxially with, and rigidly coupled to, the planet gears and mesh with the ring gear. A planetary carrier holds the gearbox components in place and rotates with the transitional and planet gears as the output of the gearbox. We designed our secondary gear train of two gears—a carrier gear and a joint gear—with the joint gear's axis of rotation acting as the knee CoR (see Fig. 3b). Every gear in the gearbox has a module and a face width, which must be shared between meshing gears, e.g., the module of the sun gear must be the same as that of the planet gear. The face width of a gear directly affects the load capacity of the gear as well as its mass. With knowledge of each gear's module $m$, face width $w$, and number of teeth $N$, it is therefore possible to design an SPC-PGT and secondary gear stage that meets the desired total transmission ratio.

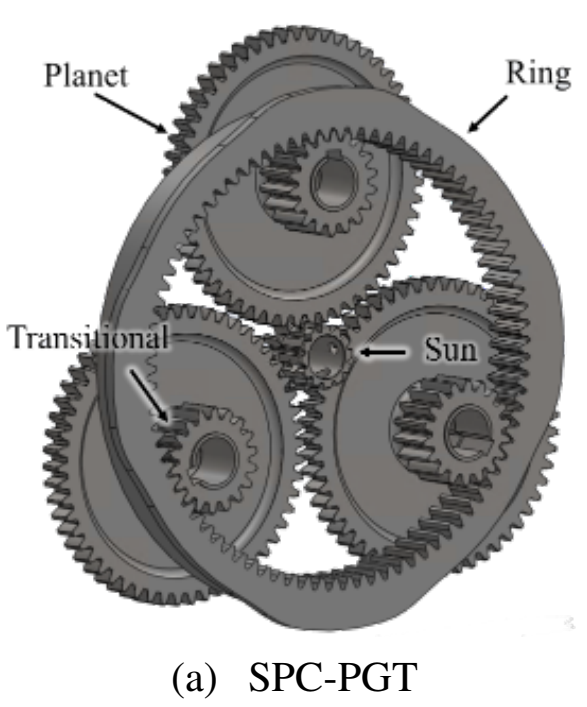

(a) SPC-PGT

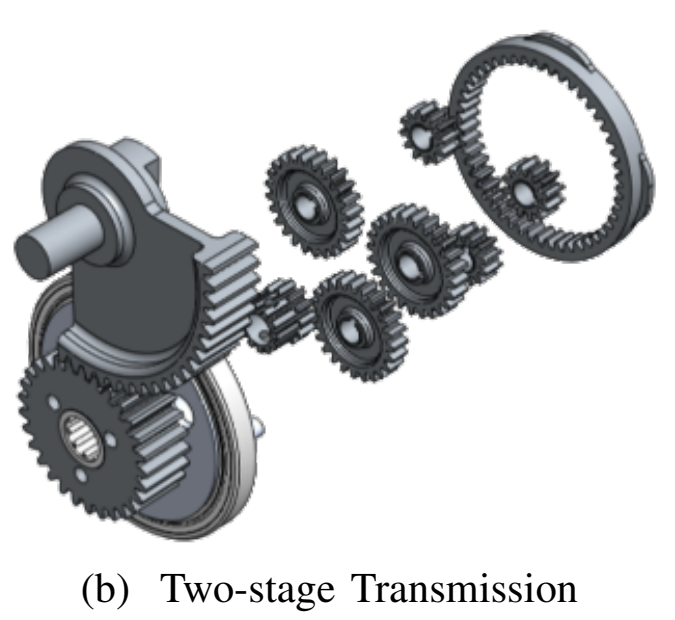

(b) Two-stage Transmission

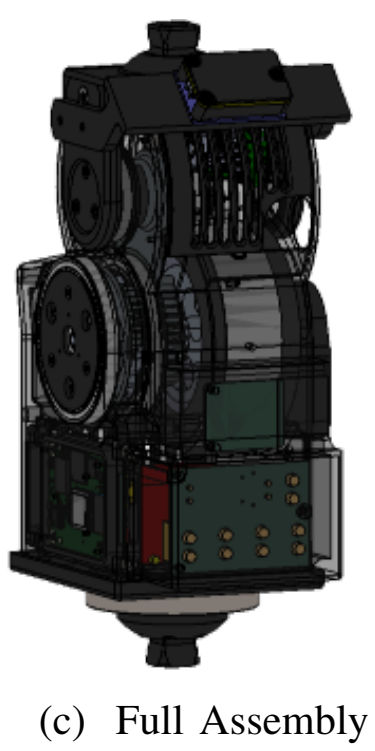

(c) Full Assembly

Fig. 3. CAD models of (a) an example SPC-PGT from [37] and (b) the two-stage, 18:1 transmission from this work, the first stage of which consists of the SPC-PGT. A transparent render of the knee prosthesis showing the internal electronic and mechanical components and their packaging is shown in (c). Regarding the SPC-PGT, the ring gear is held fixed and the sun gear acts as the input to the gear set. A carrier attached to all three transitional gears acts as the output of the first stage gear set. In the second stage, a carrier gear attached directly to the first-stage carrier meshes with the final output joint gear.

Because of the various combinations of gears that can achieve the same $i_{\mathrm{TR}}$ in a two-stage gearbox, we created a multi-objective genetic algorithm-based design optimization that finds the optimal gear design parameters for both stages, matching our desired total transmission ratio $i_{\mathrm{TR}}$ = 18:1 while minimizing the gearbox mass $M_{\mathrm{TR}}$. This optimization also minimizes the radius of the output gear at the knee CoR $R_{\mathrm{j}}$, directly impacting the minimum distance from the prosthetic knee's CoR to the residual limb. Inspired by the optimization approach for bipedal robot leg design in [44], we define the desired transmission ratio of our gearbox as $\hat{i}_{\mathrm{TR}} = \hat{i}_1 \cdot \hat{i}_2$ where

$$\hat{i}_1 = \frac{(P_{\mathrm{s}} + P_{\mathrm{p}})(P_{\mathrm{t}} + P_{\mathrm{p}})}{P_{\mathrm{s}} P_{\mathrm{t}}}, \quad \hat{i}_2 = P_{\mathrm{j}}/P_{\mathrm{c}}. \tag{1}$$

Here, $P_{\bullet}$ denotes the pitch radius of the sun (s), planet (p), transitional (t), ring (r), carrier (c), and joint (j) gears that make up both stages of our transmission. Knowing the pitch radius of our joint gear, we define the outer radius of our knee joint $R_{\mathrm{j}} = P_{\mathrm{j}} + m_{\mathrm{cj}}$ where $m_{\mathrm{cj}}$ denotes the shared module of the carrier and joint gears. The mass of the transmission $M_{\mathrm{TR}}$ is calculated from the geometry of each gear in the gearbox and the density of the gear material. We then define our multi-objective optimization as

$$X = \arg\min \left[ R_{\mathrm{j}} \quad M_{\mathrm{TR}} \quad \| \, i_{\mathrm{TR}} - \hat{i}_{\mathrm{TR}} \, \| \right], \text{ for} \tag{2}$$

$$X = \left[ m_{\mathrm{sp}}, m_{\mathrm{tr}}, \alpha_{\mathrm{sp}}, \alpha_{\mathrm{tr}}, w_{\mathrm{sp}}, w_{\mathrm{tr}}, N_{\mathrm{s}}, N_{\mathrm{p}}, N_{\mathrm{t}}, m_{\mathrm{cj}}, N_{\mathrm{c}}, w_{\mathrm{cj}}, \alpha_{\mathrm{cj}} \right]^T$$

where helix angle $\alpha_{\bullet}$ determines whether the gears in contact, i.e., the sun and planet gears (sp), are spur ($\alpha_{\bullet} = 0$) or helical ($0 < \alpha_{\bullet} \leq 30°$) gears. Helical gears can reduce the noise of the gear set, but slightly decrease efficiency and produce axial thrust that must be accounted for in bearing specification.

To avoid solutions unable to handle biomimetic loads, we utilized the Lewis Factor Equation for gear tooth stress to calculate stress on each gear given a biomimetic torque. We constrained the maximum gear tooth stress based on the material's tensile yield strength and a desired factor of safety (FOS) of 1.25 to avoid plastic deformation. We constrained the transmission ratio of the SPC-PGT to be greater than or equal to 5:1 and the transmission ratio of our second stage to be greater than or equal to 1.1:1 to ensure the second stage is providing mechanical advantage. We also constrained the distance between the carrier gear CoR and the joint gear CoR to be greater than the outermost diameter of the SPC-PGT to minimize the packaging size of the actuator. For our decision variables, we provided upper and lower bounds that followed traditional gear sizing limits from stock gear manufacturers.

To run our optimization, we provided the desired transmission ratio (18:1), desired peak joint torque (145 Nm), and the chosen gear material's tensile yield strength (1.36 GPa). The outputs of this optimization were then fed into a design table that generated CAD for each gear to be validated under biomimetic loads with finite element analysis (FEA). Further material reduction was performed iteratively in CAD and reevaluated using this FEA approach.

An exploded view of the final two-stage gearbox is shown in Fig. 3b, consisting of a 9:1 SPC-PGT and a 2:1 secondary gear train mapping the SPC-PGT carrier output to the joint (see Table SI for the final gear parameters used in this transmission design). For the gear material we selected tempered 420 stainless steel due to its high yield strength, permitting more aggressive material removal and narrower face widths for a lighter and more compact gearbox assembly compared to our previous generation knee using 304 stainless steel [37]. Ultimately, our final transmission was configured entirely with spur gears ($\alpha_{\bullet} = 0$) to reduce manufacturing complexity, given the specialized machining required for our custom gear sizes and features. The final gearbox is designed to handle a peak torque of 145 Nm with our desired FOS of 1.25.

### D. Electronics

To support high-fidelity control and modular sensor integration, we developed a custom electrical architecture centered around the Raspberry Pi Compute Module 5 (CM5; Raspberry Pi Foundation, Cambridge, England), as depicted in Fig. S1. The CM5 integrates with a custom interface board capable of various communication protocols (CAN, UART, I2C, and SPI), designed in collaboration with the Open-Source Leg project [45]. Using CAN communication with the motor driver simplified the wiring harness compared to our previous leg design [37]. The knee actuator has a single E5 optical quadrature encoder (US Digital, Vancouver, WA, USA) with 4096 cycles per revolution. This encoder is fixed to the motor shaft, and encoder data is relayed to the motor driver, which computes the motor's position and velocity. The motor position and velocity are divided by the transmission ratio of our actuator to calculate the knee's output position and velocity.

The device also incorporates a comprehensive suite of auxiliary sensors alongside a modular power distribution system. To measure ground reaction forces and moments, determine foot contact, and estimate the center of pressure, we mounted the M3564F 6-axis load cell (Sunrise Instruments, Nanning, China) at the distal base of the knee prosthesis. A custom load cell amplifier chip and ADC convert the analog signal of the load cell to a digital signal read by the CM5 using SPI. To measure thigh kinematics, the MicroStrain 3DM-CX5-IMU (HBK, Williston, VT, USA) is mounted to the knee prosthesis, communicating with the CM5 over UART. The wiring harness and electrical system can either be housed cleanly inside the knee structure or mounted externally for ease of hardware debugging and new sensor integration. We power the system with four interchangeable 450 mAh TP450-3S LiPo batteries (Thunder Power RC, Las Vegas, NV, USA) utilizing a custom power distribution board to send power from the batteries to the CM5, motor driver, and accessories.

### E. Mechanical Structure

The main mechanical structure of the knee prosthesis comprises two clamshell housings made of 7075-T6 aluminum that are fastened together to hold the motor, transmission, wiring harness, and batteries, inspired by previous devices such as the Open-Source Leg [45]. The motor is press-fit and bonded into the housing with thermally conductive epoxy to ensure direct contact between the potted windings and the housing, increasing thermal performance while keeping the housing temperature safe to touch [46]. A base plate made of 7075-T6 aluminum is attached to the bottom of both housings, further securing them. The base plate can be modified to fit a variety of pylon adapters and mounts depending on the needs of the user or researcher. The default configuration of this base plate includes mounting points for our 6-axis load cell and a distal pyramid adapter attached to the load cell's output. All moving components are packaged within this housing design, with the exception of the knee joint's proximal pyramid output. This output component, made of anodized 7075-T6 aluminum, can be redesigned or reconfigured with different socket adapters or different materials (e.g., titanium). Interchangeable 3D-printed thermoplastic polyurethane (TPU) bumpers can be placed on the knee output arm to modify the device's range of motion and reduce acoustic noise during ambulation. Without bumpers, the knee's range of motion is from -5° hyperextension to 105° flexion. In the default configuration, both the proximal output pyramid and distal pyramid adapter are aligned with the center of the device measured from one side of the housing to the other in the coronal plane. This allows the device to be worn on either the right or left side of the body. With adjustments to the base plate and output pyramid, the device can be easily individualized to the geometry of the user.

We performed FEA with Ansys Mechanical (Ansys, Canonsburg, PA, USA) to assess both structural rigidity and thermal performance of the housing under load. We designed the housing to make the components as lightweight as possible while still meeting our requirements, including the structural loads of a 116 kg user with a FoS of 3. To estimate thermal performance under load, we utilized a similar FEA setup as in Supplemental Section SIII-B2, replacing the stock heatsink with the full housing assembly. Under 600 continuous stair ascent strides with a conservative heat transfer coefficient for ambient air ($h = 10$ W/(m$^2$·K)), we estimated that the housing would reach a maximum temperature of $44.5°$ C, less than the safe-touch limit of $48°$ C [46]. Due to its reconfigurable nature, the base plate was not included in the FEA simulation. We reasonably assume that this added component may improve real-world thermal performance as the added surface area, mass, and direct contact with both sides of the housings would provide more conduction and convection.

To further support a modular setup, the internal cavity includes multiple mounting points for the electronics and wiring harness, with additional slots and features for securing wiring with Velcro or zip ties. The CM5 and motor driver are mounted directly to the frame, allowing for passive cooling of the electronics through heat transfer with the frame. The internal cavity is designed to be accessible through 3D-printed covers that are held in place with small neodymium magnets allowing for maintenance of the wiring harness and internal electrical components without requiring full disassembly of the housing. These covers can also be redesigned to allow for wiring pass-throughs and other functional design changes. Finally, the bottom cavity of the device houses the onboard batteries and the power distribution board.

## III. Experimental Validation

To evaluate the proposed knee prosthesis and establish performance benchmarks against the state-of-the-art, we conducted both benchtop characterization and in-vivo amputee experiments. Benchtop experiments quantify the intrinsic capabilities of the QDD actuator, whereas the amputee experiments validate device function under real-world loading conditions, including both high velocity and high torque regimes during level-ground walking and sit-stand transfers.

### A. Benchtop Experiments

We conducted a suite of benchtop experiments to characterize the physical capabilities of the QDD actuator and validate our baseline controller. These evaluations focused on backdrive torque, closed-loop position tracking, open-loop torque tracking, and peak torque capacity. These tests confirm our core design hypothesis: the high mechanical transparency of the QDD paradigm allows for highly accurate position and torque control using only a motor encoder, obviating the need for output-facing sensors or modeling complex, non-linear actuator dynamics.

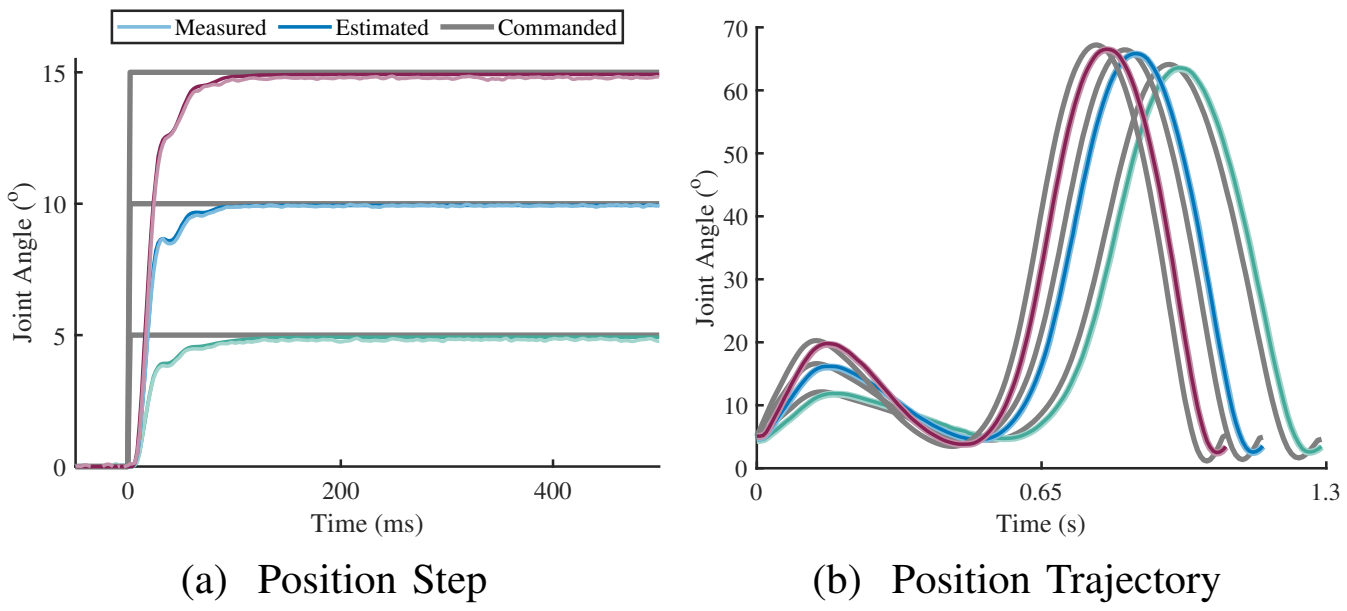

Fig. 4. Benchtop plots of (a) position step and (b) position tracking tests comparing commanded, estimated, and externally measured (ground-truth) joint angles. Teal, indigo, and purple curves represent (a) position step responses of 5°, 10°, and 15°, respectively, and (b) position trajectories for walking speeds of 0.8, 1.0, and 1.2 m/s, respectively. Commanded joint angles are shown in gray, estimated joint angles are solid colors, and measured joint angles are semi-transparent colors (closely overlapping with solid).

*1) Backdrive Torque:* To quantify the torque required at the output arm of the knee actuator to backdrive the motor rotor, we rigidly mounted the knee to a benchtop through the base plate and attached the inner diameter of the M3564F 6-axis load cell in series with the output arm's axis of rotation through a custom coupler. A lever arm was attached to the outer radius of the load cell, allowing us to measure the torque required to induce a change in joint angle. Ten static backdrive torque measurements were taken—five in flexion and five in extension—with the output arm positioned vertically to avoid gravitational assistance in backdriving the actuator. The average static backdrive torque was approximately 0.98 Nm, matching our previous generation knee [37].

*2) Closed-Loop Position Tracking:* To assess the ability of our device to perform closed-loop position tracking, we performed two different benchtop tests, a step response test and an able-bodied trajectory tracking test. A PD controller was tuned using the Ziegler-Nichols method as a starting point and then further hand-tuned. Our proportional gain, $k_\text{p}$, was 300 Nm/rad (5.236 Nm/deg), and our derivative gain, $k_\text{d}$, was 7 Nms/rad (0.122 Nms/deg). For both tests, the knee prosthesis was rigidly mounted at its base. For the AB joint trajectory tracking test, a MicroStrain 3DM-CX5-IMU was mounted on the upper link to directly measure the joint angle, i.e., the transmission output rather than the input measured by the encoder. Because the IMU's onboard gyroscope filtering algorithm attenuates rapid transients, the high-acceleration position step responses were measured with motion capture (Vicon Ltd. Oxford, UK).

The step response tests followed a similar protocol as outlined in [30]. We commanded the knee to track steps of 5°, 10°, and 15°, performing each test five times with the knee output starting at the hyperextension hardstop. Fig. 4a shows the average step responses, comparing the commanded position to the estimated joint position (via the motor encoder) and the measured joint position (via motion capture). The estimated rise times are between 33 and 48 ms, while the measured rise times are between 35 and 54 ms, demonstrating a highly responsive control loop. The average RMSE between the estimated and measured knee angles across all step tests is 0.10°, demonstrating highly accurate onboard state estimation.

For the trajectory tracking tests, we commanded the knee to track AB walking trajectories at 0.8, 1.0, and 1.2 m/s, five times each. Fig. 4b shows the average commanded, estimated, and measured joint positions for each walking speed. From lowest to highest walking speed, the estimated tracking RMSE was 3.59°, 4.19°, and 4.65°, while the measured tracking RMSE was 3.65°, 4.18°, and 4.64°. Correcting for phase lag, the estimated tracking RMSE reduces to 0.34°, 0.42°, and 0.66°. Comparing measured and estimated joint positions yields an average RMSE of 0.24° across all tasks and trials. From these results, we can conclude that the estimated joint position calculated from motor position is sufficiently accurate for high-fidelity position control.

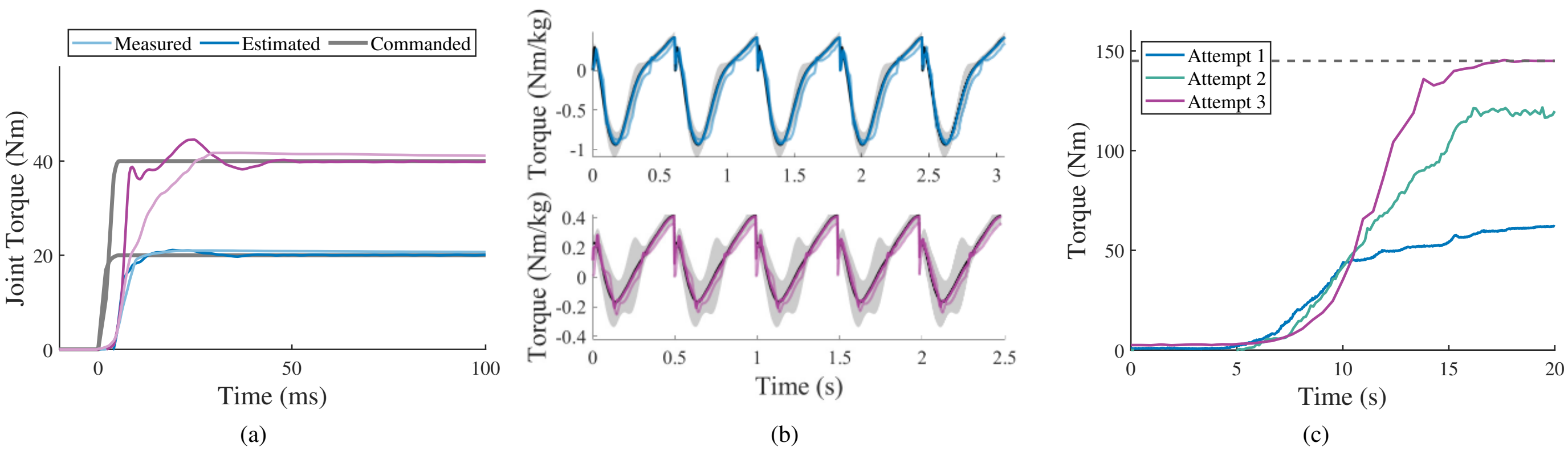

Fig. 5. Benchtop plots of (a) torque step responses, (b) stair and walk torque trajectory tracking, and (c) peak torque tests. In (a), the commanded joint torque is denoted in gray, the joint torque estimated from the motor driver's q-axis current is denoted in solid colors (indigo for 20 Nm, purple for 40 Nm), and the torque measured by the external load cell is denoted by semi-transparent colors. In (b), torque trajectories are given for a 95 kg participant during level-ground walking at 1.0 m/s (bottom) and stair ascent on a 30° incline (top). The commanded normative AB joint torque is denoted with a black dashed line, and the surrounding light gray region indicates ±1 standard deviation of the normative AB reference data. Solid colors represent estimated torque values and semi-transparent colors represent measured values (indigo for stairs and purple for level-ground walking). Sub-figure (c) shows increasing manually applied loads to reach 145 Nm at the knee joint.

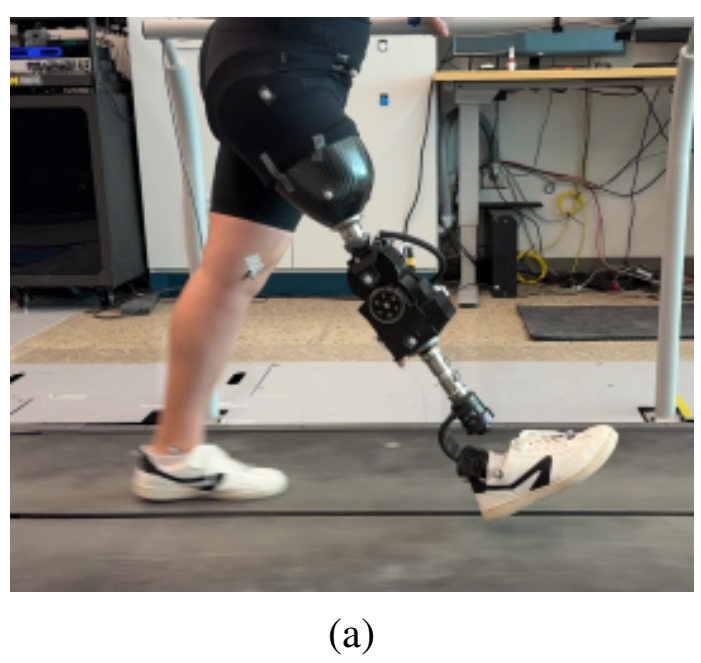

(a)

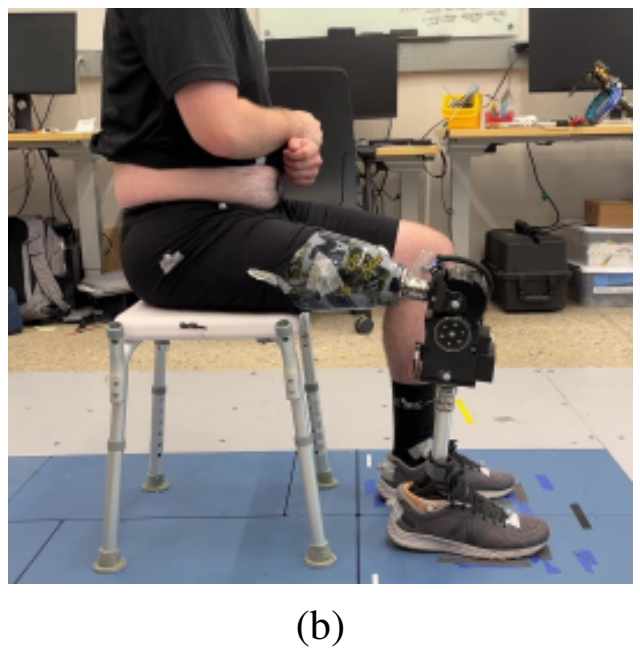

(b)

Fig. 6. Pictures of two amputee participants wearing the QDD knee prosthesis during (a) level-ground walking and (b) sit-stand experiments.

*3) Open-Loop Torque Tracking:* To evaluate open-loop torque control (tracking based solely on motor current), the knee prosthesis was rigidly secured to the testbench at its distal base with an external load cell mounted in series at the fixed proximal output. Using this configuration, we evaluated both torque step responses (20 Nm and 40 Nm) and able-bodied (AB) torque trajectories corresponding to level-ground walking (1.0 m/s) and stair ascent (30° incline) for a 95 kg user [21], [43], following validation protocols adapted from [30]. Each torque step and trajectory was commanded five times.

Fig. 5a shows the average commanded, estimated, and measured data for the 20 and 40 Nm torque steps. Across both torque levels, the rise times measured between 10.19 and 17.40 ms, matching state-of-the-art performance in [30]. The results of our AB torque trajectory tracking tests are shown in Fig. 5b for commanded stair ascent and walking torques. Across stair ascent trials, the average peak extension torque was within 0.05 Nm/kg of the commanded peak torque. To contextualize these tracking errors against biological variability, this discrepancy is less than one-third of the natural AB standard deviation ($\pm$0.17 Nm/kg) observed in maximum knee extension torque. Similarly, the average peak extension torque for walking was within 0.07 Nm/kg of the command, which is likewise smaller than the normative biological standard deviation of $\pm$0.12 Nm/kg.

*4) Peak Torque Test:* To evaluate the device's ability to output the high torques necessary for tasks like stair climbing and to benchmark against the state-of-the-art [31], we performed a 145 Nm peak torque test. During this experiment, the knee actuator was fixed to the benchtop, with an extended pylon attached to the output pyramid. While the previously described position controller maintained a fixed angle, three incremental loads were manually applied through the pylon until the target of 145 Nm was reached (Fig. 5c).

### *B. Amputee Participant Experiments*

Device function was evaluated during level-ground walking at three common speeds and during sit-stand transitions with two K4 (P1 and P2) and one K3 (P3) TF amputee participants (see Supplemental Table SII for participant details). The following experimental protocol was approved by the University of Michigan Institutional Review Board (HUM00230065).

*1) Experimental Methods:* To perform level-ground and sit-stand activities, we implemented our previously developed Hybrid Kinematic Impedance Control (HKIC) framework from [47], [48]. The stance impedance model, swing kinematic model, and respective control parameters were left unchanged from their implementation on our previous-generation knee-ankle prosthesis. While the lack of an actuated ankle in the presented prosthesis impacts this biomimetic knee controller [14], the goal of the following tests was to assess the device performance under realistic loading conditions rather than to evaluate the controller design.

For these experiments, the knee prosthesis was configured as shown in Fig. 6, powered by four onboard 450 mAh 3S LiPo batteries (Thunder Power RC, Las Vegas, NV, USA). Utilizing their daily-use prosthetic foot, each participant was fitted to the knee prosthesis with a custom pylon tube connected via tube clamps to the knee's distal pyramid and the foot's proximal pyramid, ensuring level standing and walking. Participants wore a matching pair of their own shoes on both the prosthetic and contralateral sides.

The experimental protocol centered around a primary 4-hour data collection session, with total participation time tailored to each individual's prior experience with powered prostheses. Because P1 and P3 had prior experience with our HKIC architecture on other systems, both completed their acclimation and testing within this single session. In contrast, P2 had no prior experience with powered knees and therefore completed an additional acclimation session a week before their experimental data collection. All participants were fitted by a licensed prosthetist and given time to acclimate to the device. Before the start of data collection, we allowed all participants to perform self-paced walking and sit-stand transitions utilizing parallel bars until they felt comfortable with the device fitting and control behavior. A safety harness and hand rails were available at all times throughout acclimation and data collection.

During each data collection session, participants performed level-ground walking and sit-stand transitions. For the walking experiments, participants walked on a split-belt instrumented treadmill (Bertec, Columbus, OH, USA) at speeds of 0.8, 1.0, and 1.2 m/s. Each speed condition consisted of a single trial that lasted 90 seconds, where the first and last five strides of each trial were discarded to assess device performance during steady-state walking. We then asked the participants to perform 10 sit-stand-sit repetitions starting from standing, utilizing a stool.

To measure real-world acoustic noise levels, P1 performed an additional 20 trials of level-ground walking at 1.0 m/s on flat ground while a PCE-322A sound level meter (PCE Instruments, Florida, USA) recorded A-weighted sound pressure level. The meter was placed at the height of the device's actuator, 1.5 m away from the center of the walking path. Each recording consisted of a single steady-state stride taken directly in front of and perpendicular to the meter. Ten strides were performed with the prosthetic side closest to the meter, and ten were performed with the participant's sound side closest. The participant was given a walking start to achieve steady-state gait in front of the sensor. For comparison, the same sound measurement protocol was performed with the Össur Power Knee running the same HKIC framework [14].

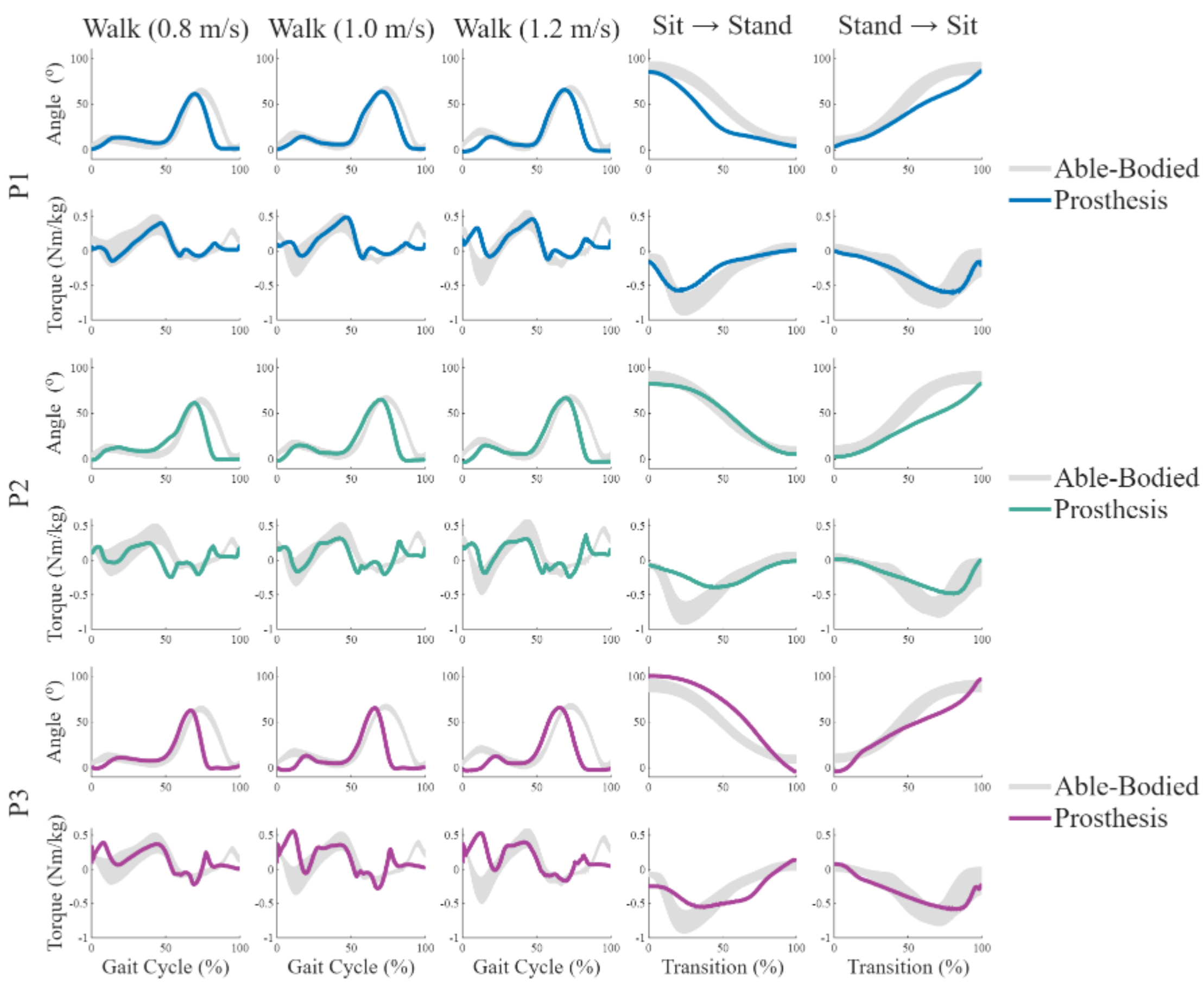

Fig. 7. Mean prosthetic kinematics and kinetics for participant P1 (indigo), P2 (teal), and P3 (purple) during level-ground walking at three speeds (0.8, 1.0, and 1.2 m/s) and during sit-stand transitions. For comparison, AB averages $\pm 1$ standard deviation are shown in gray [43].

*2) Level-Ground Walking Performance:* During level-ground walking, the prosthesis demonstrated biomimetic trends across participants and walking speeds, with peak kinematics and kinetics largely falling within one standard deviation of able-bodied (AB) reference data [43]. Columns one through three of Fig. 7 show these average knee angles and torques recorded by onboard sensors over the three walking speeds, comparing them to AB references.

The prosthetic biomechanics slightly led the AB reference for all participants, reflecting the typical amputee gait compensation of shortened stance-phase duration on the prosthetic side, which shifts swing kinematics earlier in the normalized gait cycle. The phase-based walking controller [47] synchronized the knee joint patterns to this shifted stride progression. Despite this temporal difference, we observed clear biomimetic trends, such as increasing knee flexion angle with increasing walking speed. Across participants, we also observed delays in early-stance knee flexion angle and knee extension torque with increasing walking speed. This behavior was most noticeable in the biomechanics of P3, who, as a K3 ambulator, was less accustomed to walking at faster speeds. Despite this delay in early-stance knee flexion, P1 and P2 achieved peak knee extension torques within one standard deviation of the AB reference at all walking speeds, while P3 achieved comparable peaks for 0.8 m/s and 1.0 m/s walking.

*3) Sit-Stand Transition Performance:* During sit-stand transitions, the prosthesis generated near-biomimetic joint torques while accommodating individual user preferences and timing variations via the phase-based controller. Columns four and five of Fig. 7 compare the average prosthesis knee kinematics and kinetics during sit-to-stand and stand-to-sit transitions against the AB reference biomechanics from [41], which were used to build the HKIC model [48]. Due to the time-invariant nature of the thigh-based phase variable, participants could indirectly control their progression, resulting in temporal differences between participants and with the AB reference. We also observed differences between participants with respect to their preferred sitting and standing knee configurations, likely due to variations in stiffness and geometry between their passive feet, participant physiology, and personal preference (see Section IV). Across participants, peak knee extension torques fell within one standard deviation of the AB reference during stand-to-sit transitions, while torque was slightly reduced during sit-to-stand transitions. These trends matched those observed with the same control method on the Össur Power Knee, suggesting this discrepancy may be impacted by the lack of a powered ankle [14].

*4) Acoustic Noise Results:* The presented prosthetic knee demonstrated a substantial reduction in acoustic emissions compared to the commercial benchmark—the Össur Power Knee running the same HKIC controller—during level-ground walking at 1.0 m/s. The presented prosthesis exhibited an average acoustic noise level of 48.54 dB, approximately 3.9 dB quieter than the Power Knee (52.33 dB) throughout the entire gait cycle. Furthermore, the average peak noise level of the presented prosthesis (51.65 dB) was approximately 4.84 dB quieter than the Power Knee (57.17 dB) across the ten trials. Due to the logarithmic nature of decibel measurements, these deltas represent substantial differences in acoustic sound power. Notably, the ambient noise in the room was approximately 39 dB on average throughout the testing period, which may have artificially inflated these averages by masking lower device emissions throughout portions of the stride.

## IV. Discussion

### A. Device Capabilities Compared to State-of-the-Art

This work demonstrates that a low-impedance, QDD actuator can be packaged in a lightweight and compact platform, weighing $\sim$1 kg less than the state-of-the-art QDD knee [37]. The presented design is $\sim$100 g lighter and 6.3 cm shorter than the commercial Össur Power Knee despite being capable of almost twice the peak joint torque (145 vs. 80 Nm), confirming the feasibility of the QDD architecture for future clinical and commercial use. While the presented design is 700 g heavier than the Utah Direct Ball Screw Drive Knee [31], our design can provide 145 Nm active torque across its entire range of knee flexion (up to 105°), whereas the Utah Knee's active torque is limited to 115 Nm and degrades at high flexion angles due to the transmission's kinematic singularity at 88°.

Our lightweight 18:1 actuator also offers output impedance properties that are competitive with state-of-the-art low-impedance designs. The presented static backdrive torque ($<$ 1 Nm) matches our previous-generation QDD knee [37], and the presented actuator reflected inertia (0.037 kg·m²) is 33.9% smaller than this prior design (0.056 kg·m²). While the latest-generation Utah Knee [31] has a slightly lower reflected inertia (0.011–0.030 kg·m²), our reflected inertia is constant across all loading profiles and less than 10% of the pendulum inertia of the biological shank and foot ($0.1 \cdot 0.38$ kg·m² [20]). This feature of our design enabled accurate impedance and position control without an output torque sensor or encoder.

Throughout the walking and sit-stand experiments, the device provided comparable knee biomechanics to reference AB kinematics [41], [43] across a heterogeneous cohort of three participants varying in height (1.60–1.75 m), weight (68.40–89.20 kg), sex, mobility level (K3–K4), and underlying clinical presentation. Due to compromised contralateral limb strength, P3 exhibited greater reliance on the powered prosthesis for loading and balance during sit-stand transitions, reflected by increased knee flexion while sitting, hyper-extension of the knee while standing, and a longer high-torque regime observed during sit-to-stand kinetics (see Fig. 7). The ability to provide high torque at high knee flexion angles is an advantage of the QDD design compared to the kinematic linkage-based transmission of [31], which cannot provide active torque assistance at its singular configuration at 88°. The ability of the presented device to accommodate user differences highlights its potential to adapt to the varied physiology and behavioral traits of the broader transfemoral amputee population.

Similar to our previous device, the knee prosthesis emits low acoustic noise under load, falling under the Össur Power Knee in both average and peak sound power metrics under identical control and task conditions. To provide qualitative context, the sound level of the presented prosthesis (48.54 dB average, 51.65 dB peak at $\sim$1.5 m) resembles a quiet library (40 dB to 50 dB [49]), rising to the hum of a refrigerator condenser (50 dB) at its peak. While A-weighted sound power level does not fully capture perceived loudness or annoyance, the noise differences between devices including pitch were highly perceivable by the users. While not a controlled comparison, our observed peak noise emission is also lower than the 53 dB peak reported for the latest generation Utah Knee in [31].

### B. Participant Feedback

Qualitative feedback from all three participants confirmed distinct advantages in the device's weight, acoustic noise/pitch, and/or stance-phase support compared to their daily-use devices and previously tested powered devices. While P2 had no prior powered device experience, they felt a minimal weight difference compared to their prescribed prosthesis while reporting reduced effort and less back strain while walking. Both P1 and P3, who had prior experience with powered devices including the Össur Power Knee, noted the QDD knee enabled better perception of the passive foot's behavior during early-to-mid stance while walking. P1 noted that this transparency gave them increased confidence throughout the stride and mirrored the proprioceptive feeling they had previously experienced only with commercial microprocessor knees.

Constructive feedback from P3, however, revealed that users with limited mobility in their intact limb require a greater range of motion during sit-stand transitions than standard AB averages. Due to reduced strength and range of motion in their intact leg, P3 relied heavily on the prosthetic side to push up from a chair. This reliance necessitated a peak knee flexion angle of almost 120°, which was prematurely blocked at 105° by the QDD knee's hardstop. This highlights a limitation of designing around average AB biomechanics, which will be addressed in subsequent iterations to account for the compensatory mechanics required by less mobile users.

### C. Limitations and Future Work

While the 2.6 kg mass is an achievement for QDD prosthesis design compared to [37] and rivals commercial devices like the Össur Power Knee, it is heavier than the $\sim$1.9 kg Utah Knee [31]. Future work will explore removing the current commercial 6-axis load cell to save approximately 0.20 kg. This component could be replaced with a lightweight, custom, strain-gauge-based sensor similar to the Power Knee, or by leveraging alternative ground reaction force detection methods like proprioceptive force estimation [50].

Managing the trade-offs between system mass, electrical efficiency, and operational endurance remains a critical design challenge for fully untethered, high-torque prostheses. Currently, the interchangeable LiPo batteries contribute a substantial portion of our device's non-structural mass. Although LiPo chemistry possesses a lower specific energy than the Lithium-ion cells utilized in the Utah Knee and Össur Power Knee, it was selected due to its superior specific power, which is necessary to deliver the high transient currents required by our low-gear-ratio QDD architecture. Utilizing the thermal model (see Section SIII-B) and assuming a worst-case power draw, we estimate the device is capable of approximately 8,300 continuous walking strides at 1.0 m/s for a 102 kg user, providing under 2 h of continuous operation. While analogous continuous usage limits are not reported for the Össur Power Knee, this device is rated for 4–20 h of *intermittent* operation depending on activity level and user weight [26]. Future iterations will aim to extend our operating envelope through higher-capacity, custom-contoured LiPo packs optimized for the internal electronics compartment. Improving component thermal and electrical efficiency will also reduce parasitic power draw (e.g., using a $\sim$1 W production-grade MCU) to improve commercial viability.

While we demonstrated accurate, high-fidelity torque control through our benchtop and in-vivo evaluations, we have yet to perform ramp and stair activities with amputee participants. Consequently, future work will focus on evaluating a broader range of ADLs, requiring adaptation of our biologically-inspired HKIC control suite for standalone knee systems with non-biomimetic ankle behavior as in [51]. As the Össur Power Knee served as the primary structural and inertial benchmark for this design, we plan to perform clinical experiments directly evaluating the two devices running identical control architectures across comparative metrics such as biomimicry, thermal characteristics under load, and participant perceptions of comfort, mobility, and exertion.

Future work also includes extending this knee prosthesis design to a lightweight, QDD-powered ankle using a similar design optimization and modeling framework. This device will employ the same electronics architecture as the knee, allowing for a unified, multi-node communication bus over CAN. As a result, this knee-ankle prosthesis paradigm will be fully modular—supporting both independent and synchronized joint operation—while maintaining a lower mass profile than our previous-generation knee-ankle system [37]. Integrating both a powered knee and ankle will allow us to leverage the full potential of our HKIC control framework [52] and draw meaningful comparisons with other state-of-the-art fully-powered leg platforms [30], [45].

## V. Conclusion

In this work, we presented the design and validation of a lightweight, highly backdrivable, QDD knee prosthesis. To achieve a minimal mass and size profile, we developed a multi-stage transmission design optimization framework that streamlines the mechanical iteration process. Following hardware fabrication, we validated the device's torque and position control capabilities through benchtop experiments. Furthermore, in-vivo evaluations with three unilateral amputee participants across walking and sit-stand transitions demonstrated robust mechanical performance across a wide joint range of motion, alongside a quantifiable reduction in acoustic noise compared to the commercial Össur Power Knee [26].

The presented hardware platform supports tuning-free, biomimetic control strategies [47] via high-fidelity position and torque tracking—capabilities that remain challenging for devices operating in high- or variable-impedance regimes. While some state-of-the-art research platforms remain lighter, our QDD device retains the benefits of a constant low gear ratio (e.g., backdrivability and controllability) across knee flexion angles and loading conditions. Furthermore, when benchmarked against current commercial prostheses, the presented device maintains a commercially viable mass while offering significantly reduced mechanical complexity and quieter operation. Together, these advantages showcase the definitive promise of QDD architectures for future clinical and commercial translation.

Future work will focus on evaluating this knee prosthesis during variable locomotor tasks, such as stair ascent and descent, alongside clinical testing across a larger cohort of amputee participants. Additionally, the development of a structurally matching QDD ankle will enable a fully integrated knee-ankle platform and facilitate direct, system-level comparisons with other fully powered devices. Ultimately, this platform provides both a foundation for advanced control architectures and a benchmark for the continued evolution of highly backdrivable, QDD prosthetic systems.

## Acknowledgments

The authors thank T. Kevin Best and Curt Laubscher for assisting with experiments and our participants for their time and feedback. We thank Nikko Van Crey, Riley Pieper, Japmanjeet Singh Gill, T. Kevin Best, Senthur Raj Ayyappan, and Elliott Rouse for participating in design reviews and co-development of the Open-Source Leg code base and Elmo drive CAN library. The authors also thank Alan Tondryk, Japmanjeet Singh Gill, and Elliott Rouse for development of the CM5 interface board and load cell amplifier. Power Knee hardware was provided by Össur hf (Reykjavík, Iceland).

# Supplementary Methods and Materials for “Design and Validation of a Lightweight, Low-Profile Powered Knee Prosthesis with Quasi-Direct Drive Actuation”

## I. Design

### A. Electronics

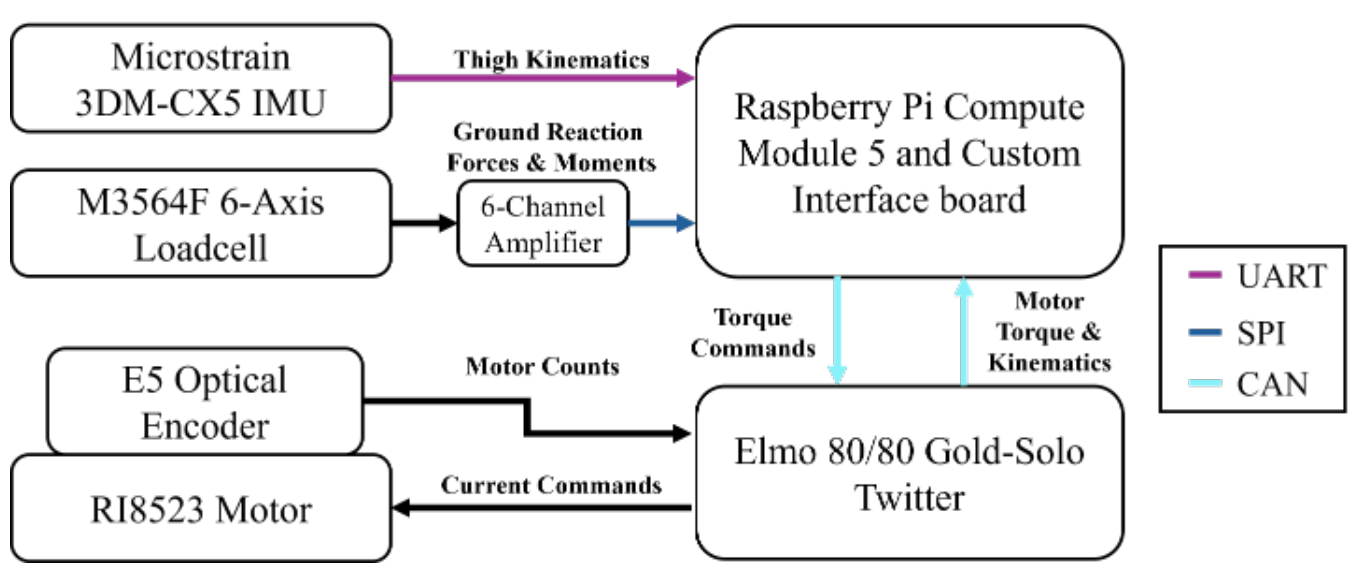

Fig. 1. Block diagram of the device’s electrical system. The direction and contents of communication between components is shown with labels and arrows indicating directionality. UART communication is colored purple, SPI communication is colored indigo, and CAN communication is colored teal.

### B. Transmission

TABLE I
Transmission Gear Parameters

| Name | # of Teeth ($N$) | Module ($m$) | Face Width ($w$) |
|---|---|---|---|
| Sun | 12 | 1 | 5 mm |
| Planets | 24 | 1 | 5 mm |
| Transitional | 12 | 1 | 6 mm |
| Ring | 84 | 1 | 6 mm |
| Carrier | 29 | 1.25 | 13.21 mm |
| Joint | 58 | 1.25 | 13.21 mm |

## II. Experimental Validation

TABLE II
Participant Information

| ID | Sex | Age (yrs) | Mass (kg) | Height (m) | Prescribed Device (knee/ankle) | K-Level |
|---|---|---|---|---|---|---|
| P1 | M | 24 | 89.20 | 1.75 | X3/Pro-flex | K4 |
| P2 | F | 30 | 68.40 | 1.60 | X3/All Pro XTS | K4 |
| P3 | M | 56 | 80.00 | 1.73 | Power Knee/Taleo | K3 |

## III. Motor Characterization and Thermal Modeling

### A. Motor Characterization

We model our motor torque $\tau_{\mathrm{m}}$ as a function of q-axis current ($i^{\mathrm{q}}$), motor rotor velocity ($\dot{\theta}_{\mathrm{m}}$), and motor rotor acceleration ($\ddot{\theta}_{\mathrm{m}}$) with the equation:

$$\tau_{\mathrm{m}} = k_{\mathrm{t}}^{\mathrm{q}} i^{\mathrm{q}} - b_{\mathrm{m}}\dot{\theta}_{\mathrm{m}} - f_{\mathrm{m}}\mathrm{sgn}(\dot{\theta}_{\mathrm{m}}) - J_{\mathrm{m}}\ddot{\theta}_{\mathrm{m}}, \tag{1}$$

where $k_{\mathrm{t}}^{\mathrm{q}}$ denotes our motor’s q-axis torque constant, $b_{\mathrm{m}}$ denotes the viscous damping coefficient, $f_{\mathrm{m}}$ denotes motor Coulomb friction magnitude, and $J_{\mathrm{m}}$ denotes motor rotor inertia. Due to our use of a frameless internal rotor motor, $J_{\mathrm{m}}$ was found by summing the inertia of the rotor provided with the motor as well as the rotor shaft we design. The rest of our motor parameters were found performing characterization tests outlined below. We utilized multiple tests and averaged the results to estimate the motor parameters. These characterization tests are based on previous works focused on the empirical characterization of motors and actuators [1].

The test bench setup for these characterization tests is shown in Fig. 2. Our test bench utilized an analog torque sensor with a max torque of 18 Nm (Futek Advanced Sensor Technologies, Irvine, CA, USA) and two optical encoders (US Digital, Vancouver, WA, USA) to measure the position of the two motors’ rotor shafts. We characterized the RI8523 detailed in the main text, a custom, frameless internal rotor motor (T-Motor, Nanchang, China). The opposing motor was the MN1005 framed exterior rotor (T-Motor, Nanchang, China). We controlled both motors utilizing a Raspberry Pi 4B (Raspberry Pi Foundation, Cambridge, United Kingdom) over CAN with 50/100 Solo Gold Twitter motor drivers (Elmo Motion Control, Petah Tikva, Israel). For all tests, we sampled at a frequency of approximately 500 Hz.

*1) Constant Current Test:* The first method we utilized to find $k_{\mathrm{t}}^{\mathrm{q}}$ required locking the motor rotor shaft so it is held at a constant position while commanding a constant current. Because the motor rotor shaft was held at a constant position, we rewrite our original torque model as

$$\tau_{\mathrm{m}} = k_{\mathrm{t}}^{\mathrm{q}} i^{\mathrm{q}} - \cancel{b_{\mathrm{m}}\dot{\theta}_{\mathrm{m}}} - \cancel{f_{\mathrm{m}}\mathrm{sgn}(\dot{\theta}_{\mathrm{m}})} - \cancel{J_{\mathrm{m}}\ddot{\theta}_{\mathrm{m}}} + \phi \tag{2}$$

where $\phi$ accounts for the torque sensor’s bias. For each trial, we commanded a range of currents from ±0-10 A, each for 4-second intervals, disregarding the first and last 0.5 seconds, allowing the system to reach a steady state. After completing 5 trials we averaged the measured current and torque at each current interval, utilizing a simple linear regression to determine our torque constant.

*2) Back-EMF Test:* Our second method for determining our motor’s torque constant involved measuring the line-to-line voltage, $V^{ll}$, across two of the motor’s three winding leads while the motor shaft rotated at a constant no-load speed. In our experimental setup, we drove the motor’s shaft with the opposing MN1005 motor at a desired velocity utilizing velocity control. For this test, we commanded the MN1005 over a range of constant velocities, from ±0-500 revolutions per minute (RPM) in increments of 100 RPM. The driving motor ran at a constant velocity for 10 seconds with the first second of each test disregarded to allow the motor to reach a steady state. We utilized a NI USB-6218 DAQ (National Instruments Corporation, Austin, TX, USA) to measure the $V^{ll}$ at each velocity interval of the driven motor. We averaged the measured driven motor velocity, $\bar{\theta}_{\mathrm{m}}$, and the measured line-to-line voltage amplitude, $\bar{V}^{ll}$, at each interval. Because the motor we characterized was delta wound, we utilized the following equation:

$$\bar{V}^{ll} = \sqrt{\frac{2}{3}} k_{\mathrm{b}}^{\mathrm{q}} \bar{\theta}_{\mathrm{m}} \tag{3}$$

where $k_{\mathrm{b}}^{\mathrm{q}}$ is the driven motor's q-axis back-emf constant. Because $k_{\mathrm{t}}^{\mathrm{q}} = k_{\mathrm{b}}^{\mathrm{q}}$, we can find our torque constant through linear regression.

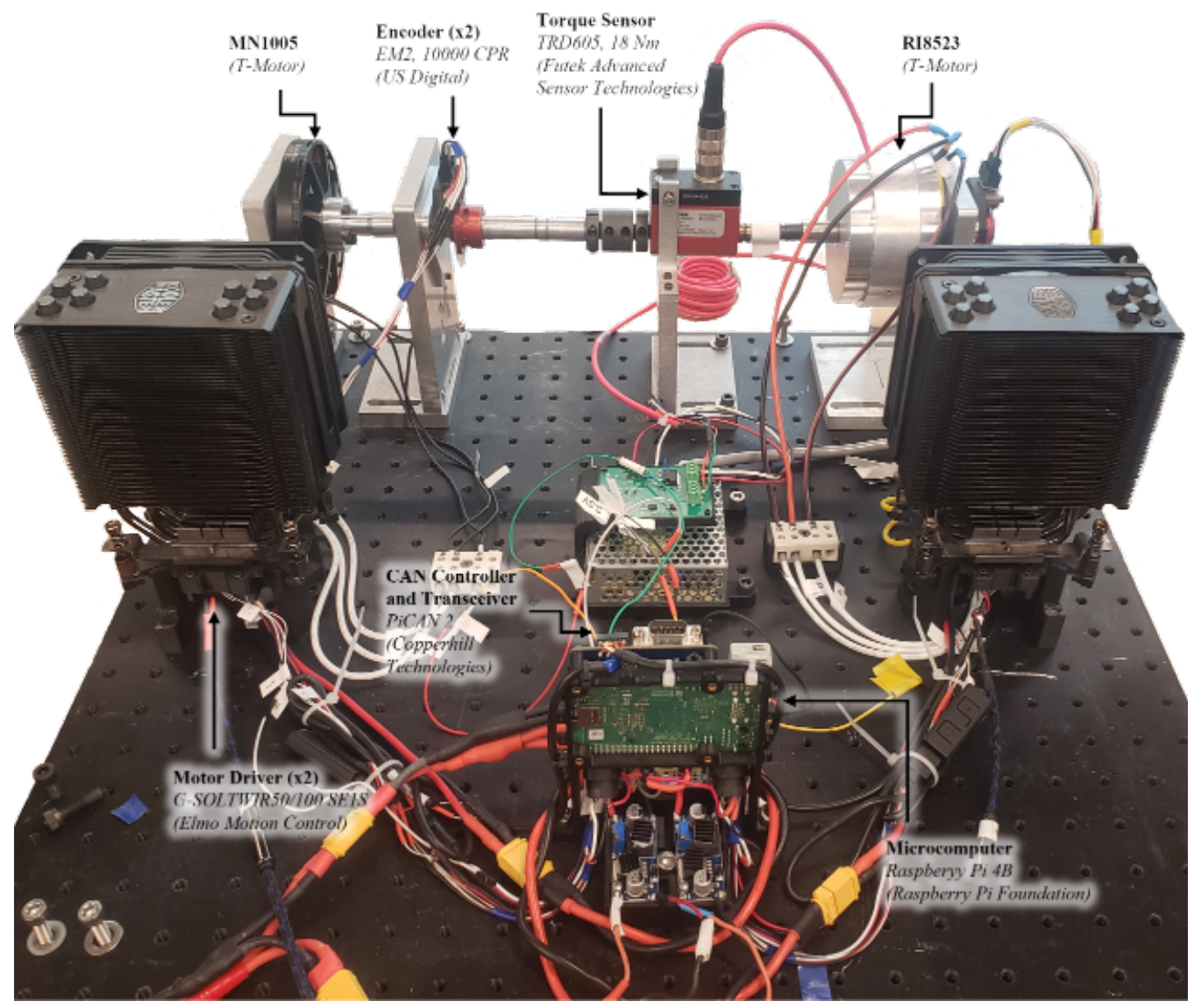

Fig. 2. A picture of our motor characterization test bench during the combined velocity and current test.

*3) Constant Velocity Test:* To determine our motor's Coulomb friction magnitude, $f_{\mathrm{m}}$, and viscous damping coefficient, $b_{\mathrm{m}}$, we commanded our motor at a constant velocity while measuring the q-axis current. We disconnected the motor's rotor shaft from the rest of the testbench, letting the motor shaft spin freely. Because the motor rotor shaft was spinning at a constant velocity with no load, we rewrite our original torque model as

$$k_{\mathrm{t}}^{\mathrm{q}} i^{\mathrm{q}} = b_{\mathrm{m}} \dot{\theta}_{\mathrm{m}} + f_{\mathrm{m}} \mathrm{sgn}(\dot{\theta}_{\mathrm{m}}) - \cancel{J_{\mathrm{m}} \ddot{\theta}_{\mathrm{m}}} - \cancel{\tau_{\mathrm{m}}}. \tag{4}$$

We conducted five trials for each velocity value ranging from ±0-500 RPM in 50 RPM intervals, disregarding the first and last second of the 4-second test to allow the rotor to reach steady-state. We then averaged the measured velocity and current at each testing interval across all five trials. Because our motor torque constant was known from our other tests, we performed a linear regression utilizing (4) to find our viscous damping coefficient and Coulomb friction magnitude.

*4) Combined Constant Current and Velocity Test:* Our final characterization test determined all three of our unknown motor parameters. For this test, we drove the motor we were characterizing at a constant velocity utilizing the MN1005. We then commanded a constant current with the driven actuator we were characterizing, measuring the resulting torque when the driven motor was at steady state. We rewrite our original torque model as

$$\tau_{\mathrm{m}} = k_{\mathrm{t}}^{\mathrm{q}} i^{\mathrm{q}} - b_{\mathrm{m}} \dot{\theta}_{\mathrm{m}} - f_{\mathrm{m}} \mathrm{sgn}(\dot{\theta}_{\mathrm{m}}) - \cancel{J_{\mathrm{m}} \ddot{\theta}_{\mathrm{m}}} + \phi, \tag{5}$$

where $\phi$ accounts for torque sensor bias. For this test, we commanded our driving motor at velocities over a range of ±0-500 RPM in 100 RPM increments and our driven motor at current values over a range of ±0-10 A in 1 A increments. For each velocity-current combination, we allow both motors to reach steady state over a 4-second interval before collecting data over a 2-second interval. For each interval, we average our measured velocity, current, and torque. After collection, we utilize the MATLAB `lsqlin` optimization function to perform a constrained least-squares regression with bounds ensuring the calculated motor parameters were all positive.

TABLE III
MOTOR CHARACTERIZATION RESULTS

| | Locked Rotor | $\bar{V}^{ll}$ | Const. Vel. | Const. Vel. & Current | Average |
|---|---|---|---|---|---|
| $\mathbf{k}_{\mathrm{t}}^{\mathrm{q}}(\frac{\mathrm{Nm}}{\mathrm{A}})$ | 0.176 | 0.168 | - | 0.181 | 0.1751 |
| $\mathbf{b}_{\mathrm{m}}(\frac{\mathrm{Nms}}{\mathrm{rad}})$ | - | - | $2.574e^{-4}$ | $3.861e^{-6}$ | $1.306e^{-4}$ |
| $\mathbf{f}_{\mathrm{m}}(\mathrm{Nm})$ | - | - | 0.035 | 0.089 | 0.062 |

*5) Results and Outcomes:* Table III shows our measured $k_{\mathrm{t}}^{\mathrm{q}}$, $b_{\mathrm{m}}$, and $f_{\mathrm{m}}$ found in each characterization test as well as the average value of each parameter across all characterization tests. This table also includes the torque constant provided in the motor's data sheet calculated in the q-axis frame. We see the benefits of performing these characterization tests as our average measured power-invariant torque constant of 0.175 Nm/A is greater than the value originally provided by the manufacturer of 0.169 Nm/A. Measuring our true power-invariant torque constant improves our understanding of the relationship between commanded q-axis current and output motor torque—in this case, less q-axis current is necessary to output our desired level of torque—and is essential for developing an accurate thermal model of the motor's behavior under load.

## B. Device Thermal Modeling

To maximize the performance of our motor we performed experiments and simulations to quantify the thermal properties of the motor, as Joule heating of a motor's windings is a major limitation of motor performance. In general, higher current commands increase the rate at which the temperatures of the windings rise. High winding temperatures can cause demagnetization and failure of the motor. By building an accurate thermal model of our motor, we can predict its thermal performance under known loads with our motor torque model developed in the previous section. We use this information to design a transmission that reduces the load on the motor and keeps the motor winding temperatures at levels we deem acceptable. We can further increase the performance of the motor by utilizing a heat sink, reducing the rate of temperature increase at the windings for the same amount of commanded current. With increased performance, we can reduce the size and mechanical advantage of the transmission, increasing backdrivability. Because the frameless motor we selected requires a frame to be mounted in our device, that frame can be modeled as a heat sink, reducing the size of the needed transmission and reducing weight.

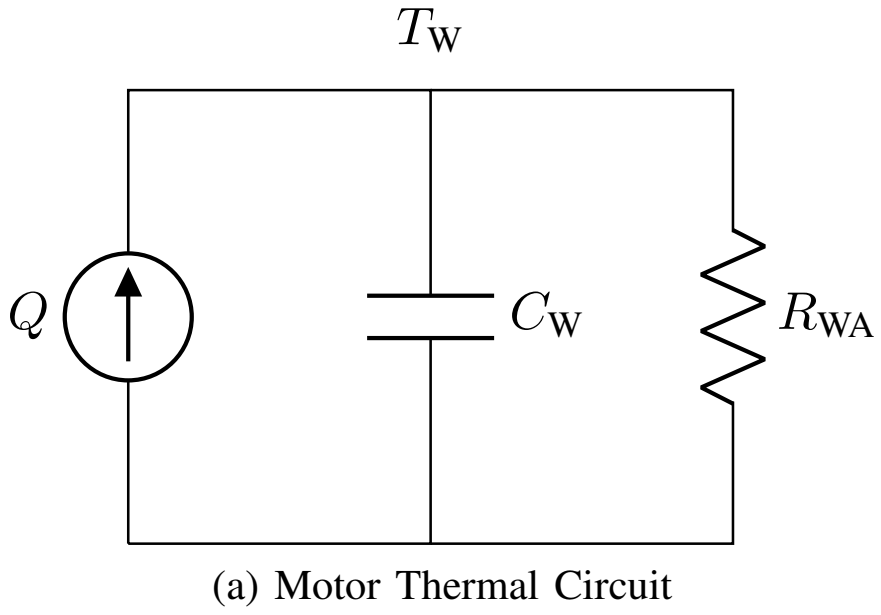

(a) Motor Thermal Circuit

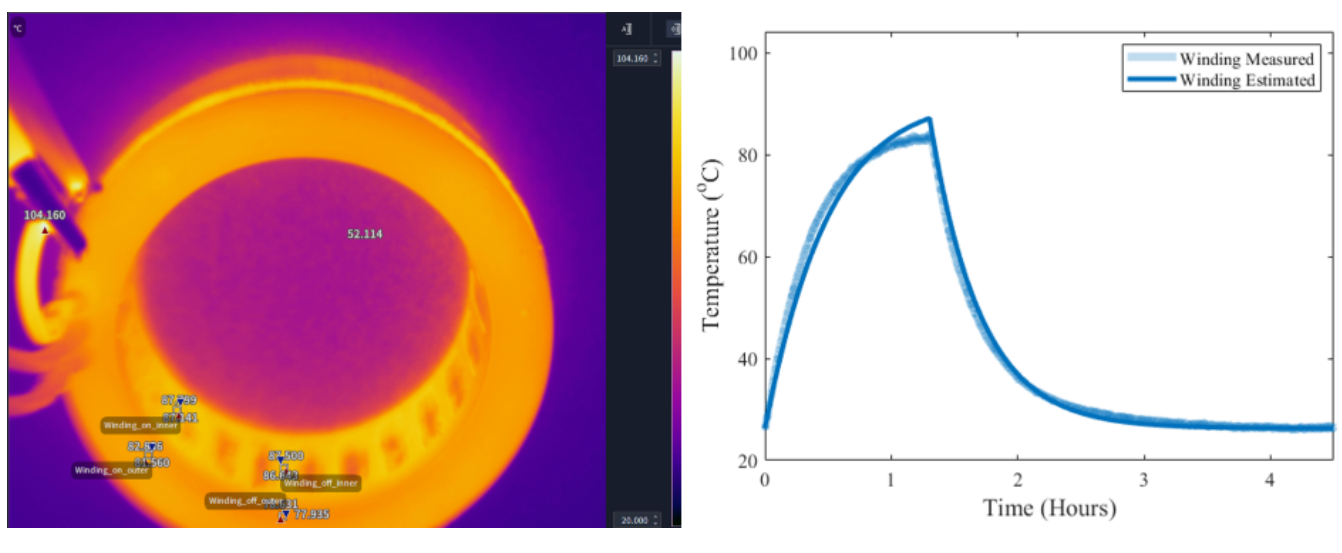

(b) Motor Thermal Modeling Results

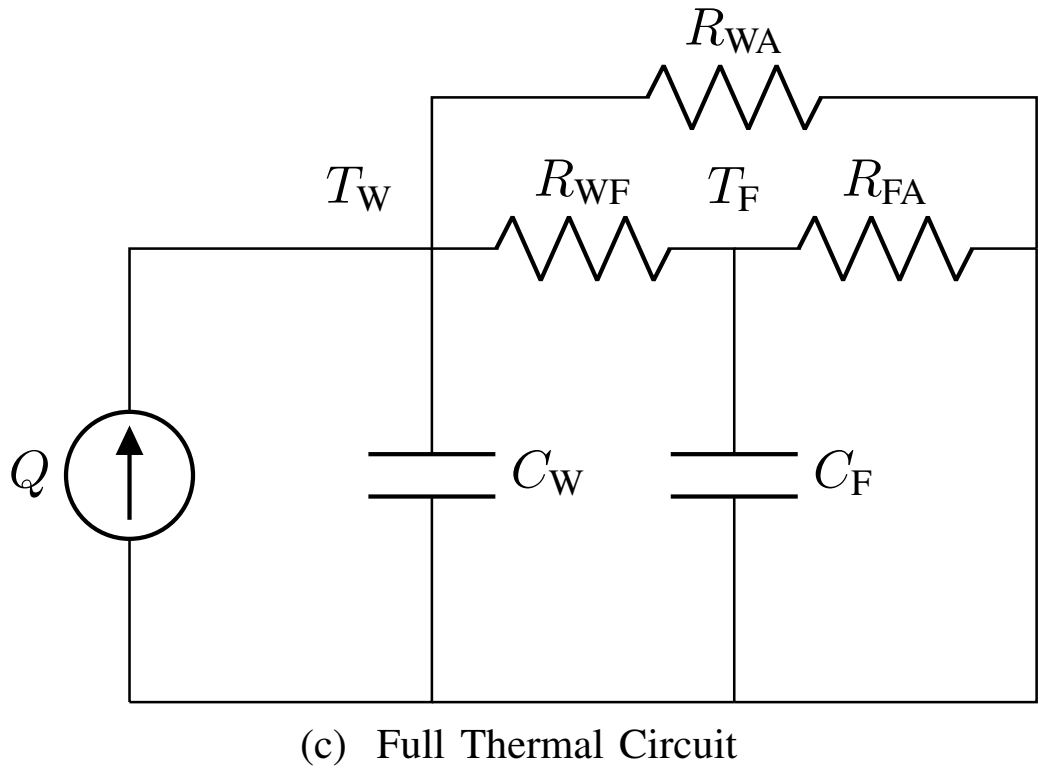

(c) Full Thermal Circuit

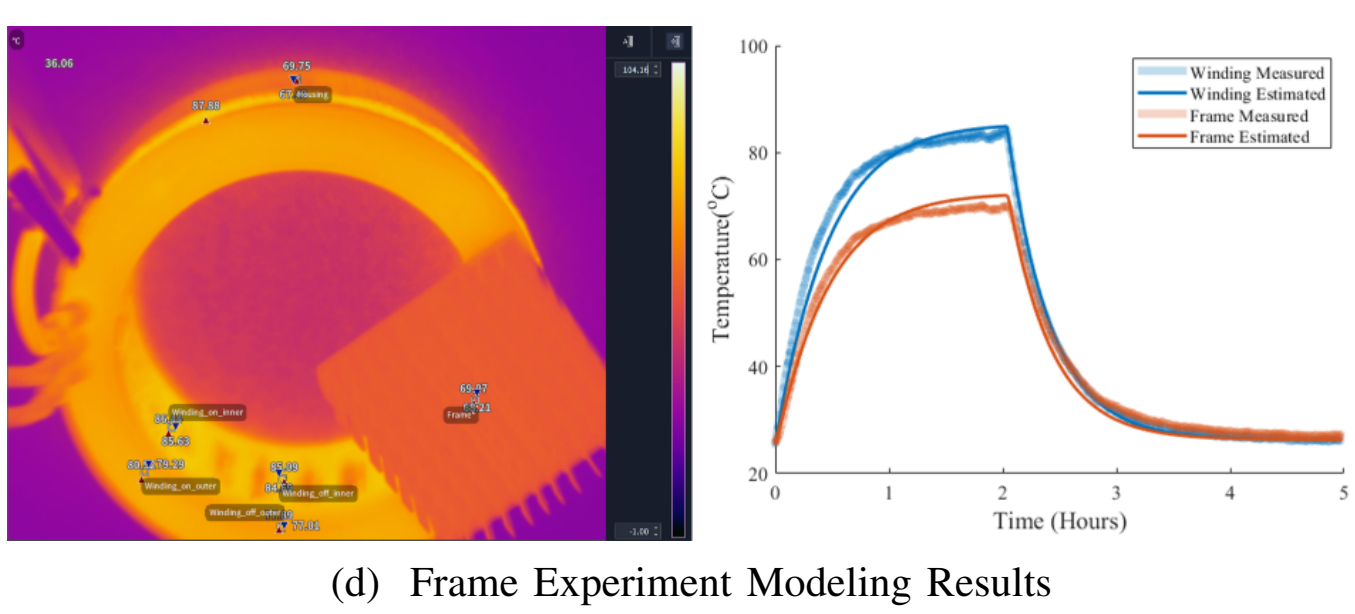

(d) Frame Experiment Modeling Results

Fig. 3. Thermal Circuit Diagrams for (a) the motor and (c) the motor frame assembly. The thermal modeling results from (b) our original motor-only experiment and (d) our motor and heatsink (denoted here as frame) experiment. The dark blue and orange lines represent our thermal model's estimated motor windings and frame temperatures. In contrast, lighter lines represent the experimentally measured temperatures from the respective experiments. Images from the thermal camera corresponding to each experiment are also shown.

A common approach to assessing motor heat sink performance is to compare the differences in motor temperature measured with a thermal camera or temperature sensor with and without the heat sink while the motor is under the same load. From these experiments, we can accurately predict heat sink performance under various loads. However, this experimental approach is not time or cost-effective as it requires long manufacturing lead times and multiple iterations to find a solution that meets all design criteria. To reduce the time and cost of this process, we utilized finite-element analysis (FEA) simulation and physical properties of the chosen frame to predict thermal performance for motor and heat sink assemblies accurately.

*1) Motor Thermal Model:* To develop an accurate thermal model for our motor, we conducted a thermal parameter identification experiment inspired by work in [1], assuming heat transfer in the motor's windings is mainly governed by conduction and convection. We modeled these modes of heat transfer by representing them in the electrical domain depicted in Fig. 3a. From this model, we write our winding temperature $T_W$ rate of change as

$$\dot{T}_W = \frac{T_A - T_W}{R_{WA} C_W} + \frac{1}{R_{WA} C_W} Q \tag{6}$$

where $Q$ is the heat flux from Joule heating of the windings, $T_A$ is ambient temperature, $R_{WA}$ is the thermal resistance between windings and ambient, and $C_W$ is the thermal capacitance of the windings.

In our parameter identification experiment, we supplied a constant current of 9 A to two of the three motor windings (active windings), measuring motor winding temperatures with an infrared thermal camera (FOTRIC, Santa Clara, CA, USA) and the voltage across the two windings with a USB DAQ (National Instruments, Austin, TX, USA), with the data sampled at 30 Hz. We allowed the motor winding temperature to reach a steady state over approximately 60 min before turning off the current source and allowing the motor to cool down to ambient temperature. From our thermal camera measurements, we calculate $T_W$ as a weighted sum of the active and inactive winding temps. Because our motor was a frameless, potted motor, we could not directly measure the winding temperatures, instead measuring the temperature of the epoxy located over the active and inactive windings. We also utilized the measured voltage and known current command to calculate the heat flux into the windings, $Q$. We used the non-linear optimization solver `fmincon` (MathWorks, Natick, MA, USA) to find the thermal model parameters that minimized the sum squared error between our measured winding temperature from our experiment and the estimated winding temperature from our thermal model. Fig. 3b shows our thermal modeling results.

*2) Frame Thermal Model:* Because our motor is frameless and potted, any frame we design will be in direct contact with the epoxied windings, resulting in conduction between the two surfaces. To accommodate these interactions and model the heat transfer occurring within the frame, we updated the electrical domain model to include frame thermal coefficients as shown in Fig. 3c. We expressed our updated model in a system of equations as

$$\begin{aligned} \dot{T}_{\mathrm{W}} &= \tfrac{T_{\mathrm{F}}-T_{\mathrm{W}}}{R_{\mathrm{WF}}C_{\mathrm{W}}} + \tfrac{T_{\mathrm{A}}-T_{\mathrm{W}}}{\hat{R}_{\mathrm{WA}}C_{\mathrm{W}}} + \tfrac{1}{C_{\mathrm{W}}}Q, \\ \dot{T}_{\mathrm{F}} &= \tfrac{T_{\mathrm{W}}-T_{\mathrm{F}}}{R_{\mathrm{WF}}C_{\mathrm{F}}} + \tfrac{T_{\mathrm{A}}-T_{\mathrm{F}}}{R_{\mathrm{FA}}C_{\mathrm{F}}}, \end{aligned} \tag{7}$$

where $C_{\mathrm{F}}$ is the thermal capacitance of the frame, $R_{\mathrm{WF}}$ is the thermal resistance between the windings and frame, $R_{\mathrm{FA}}$ is the thermal resistance between the frame and ambient, and $\hat{R}_{\mathrm{WA}}$ is the new thermal resistance between the windings and ambient. We calculated $C_{\mathrm{F}}$, $R_{\mathrm{FA}}$, using the physical and material properties of the frame, such as mass, surface area, and specific heat, utilizing known heat transfer relationships. We then determined $\hat{R}_{\mathrm{WA}}$ by scaling $R_{\mathrm{WA}}$ relative to the change in motor surface area in "contact" with ambient air. As thermal resistance due to convection is inversely proportional to surface area exposed to ambient, $\hat{R}_{\mathrm{WA}}$ increases as the frame covers more of the motor. To estimate $R_{\mathrm{WF}}$, we performed a dynamic thermal simulation on a computer-aided design (CAD) model of the frame in Ansys Mechanical to determine the relationship between motor winding and frame temperature. After calculating $T_{\mathrm{F}}$ for a given input $T_{\mathrm{W}}$ we can rewrite our second equation in (7) as

$$R_{\mathrm{WF}} = \frac{T_{\mathrm{W}} - T_{\mathrm{F}}}{C_{\mathrm{F}}} \left( \dot{T}_{\mathrm{F}} + \frac{T_{\mathrm{F}} - T_{\mathrm{A}}}{R_{\mathrm{FA}}C_{\mathrm{F}}} \right)^{-1}, \tag{8}$$

solving for our thermal resistance between the winding and the frame, $R_{\mathrm{WF}}$.

*3) Results and Outcomes:* To validate our proposed thermal modeling approach, we attached a stock heat sink (McMaster-Carr, Elmhurst, IL, USA) to our motor with thermal adhesive (Coolermaster, Taipei City, Taipei, Taiwan) and performed the same experiment used in our initial motor modeling. We again allowed the motor winding temperature to reach steady state under a 9 A load, before turning off the power source and allowing the motor windings to reach ambient temperature. We then replicated this experiment with a CAD assembly in Ansys Mechanical and performed a dynamic thermal simulation. The CAD assembly consisted of the heat sink, the thermal adhesive pad, and a thin body representing the surface of the motor attached to the heat sink. In our simulation, we assumed convection is occurring on all exposed surfaces of the heat sink and thermal adhesive pad, with a conservative convection coefficient of $h = 10\ \mathrm{W/m^2K}$. We applied a temperature source on the face of the thin motor body that is in contact with the thermal adhesive, matching the motor winding temperatures measured from our original motor characterization experiment without the heat sink. To avoid errors in simulation due to heat transfer between our temperature source, the motor body, and the ambient environment, we chose a low thermal conductivity material such as polyurethane foam and allowed no convection from any exposed surfaces of the motor body. We then calculated our thermal model parameters from the simulation data and physical properties of the heat sink.

Fig. 3d compares the measured temperature of the motor windings and heat-sink to the temperatures predicted by our thermal model. At steady-state our predicted winding temperature closely matches the measured winding temperature during our experiment, predicting a slightly larger temperature. Similarly, the predicted heat-sink temperature is larger than the actual temperature of the heat sink. Both the motor and frame temperature differences at steady-state remain within the documented error of the thermal camera of $\pm 2^\circ$ C.

From this experiment we have shown our thermal modeling approach can accurately predict motor and frame temperatures under load within an acceptable margin of error. With this knowledge, along with our accurate motor torque model, we can accurately estimate motor winding and frame temperatures when the knee actuator is under biomimetic loads. We can then design a transmission and motor frame combination that increases motor performance while reducing the weight of our actuator.